\pdfoutput=1

\documentclass[11pt]{article}

\usepackage[]{acl}

\usepackage{times}
\usepackage{latexsym}

\usepackage{graphicx}
\usepackage{booktabs}
\usepackage{tabularx}
\usepackage{amsmath, bm}
\usepackage{multirow, multicol}

\usepackage[T1]{fontenc}

\usepackage[utf8]{inputenc}

\usepackage{microtype}

\usepackage{inconsolata}

\title{Unveiling the Magic: Investigating Attention Distillation in Retrieval-augmented Generation}

\author{Zizhong Li \quad Haopeng Zhang \quad Jiawei Zhang \\
  University of California, Davis \\
  \texttt{\{zzoli, hapzhang, jiwzhang\}@ucdavis.edu} \\
  }

\begin{document}
\maketitle
\begin{abstract}
    Retrieval-augmented generation framework can address the limitations of large language models by enabling real-time knowledge updates for more accurate answers. 
    An efficient way in the training phase of retrieval-augmented models is attention distillation, which uses attention scores as a supervision signal instead of manually annotated query-document pairs. 
    Despite its growing popularity, the detailed mechanisms behind the success of attention distillation remain unexplored, particularly the specific patterns it leverages to benefit training.
    In this paper, we address this gap by conducting a comprehensive review of attention distillation workflow and identifying key factors influencing the learning quality of retrieval-augmented language models. 
    We further propose indicators for optimizing models' training methods and avoiding ineffective training.
\end{abstract}

\section{Introduction}

Large language models (LLMs) have showcased remarkable capabilities across various natural language processing tasks~ \cite{min2023recent, openai2023gpt4, ouyang2022training,zhang2023extractive,zhang2023summit}. 
However, their fixed parameters limit their ability to update knowledge in real time, resulting in the hallucination problem during generation~\cite{zhang2022improving,zhang2023language}. Additionally, these models also lack protection for sensitive training data \cite{nasr2023scalable, lin2021truthfulqa}. 
One promising method to overcome these limitations is using retrieval-augmented language models \cite{ram2023context, shi2205knn, izacard2022few, guu2020retrieval, karpukhin2020dense, khandelwal2019generalization}. 
Retrieval-augmented language models typically comprise two main components: (1) \textit{the retriever}, which selects relevant information, and (2) \textit{the reader}, incorporates this information into the generation process.
Combining these two components, retrieval-augmented language models not only improve accuracy and reliability by dynamically using external knowledge but also reduce training costs with fewer trainable parameters \cite{shi2023replug, shuster2021retrieval}. 

Various methods have been proposed to improve the coordination between the retriever and the reader \cite{karpukhin2020dense, jiang2023active}. 
Among these, attention score-based knowledge distillation has shown its effectiveness \cite{izacard2020distilling}, outperforming other established methods \cite{karpukhin2020dense, lewis2020retrieval, izacard2020leveraging} in QA tasks.
In this process, the attention scores from the reader are captured and conveyed to the retriever as the supervisory signal, enabling the retrieval model to more effectively identify information candidates that can significantly improve the language model's responses.
This efficient strategy reduces the need for manual annotation of the knowledge corpus, saving resources while achieving satisfactory results \cite{hu2023survey, wang2023survey}.

\begin{figure}[t]
    \centering
    \includegraphics[width=0.48\textwidth, height=3.5cm]{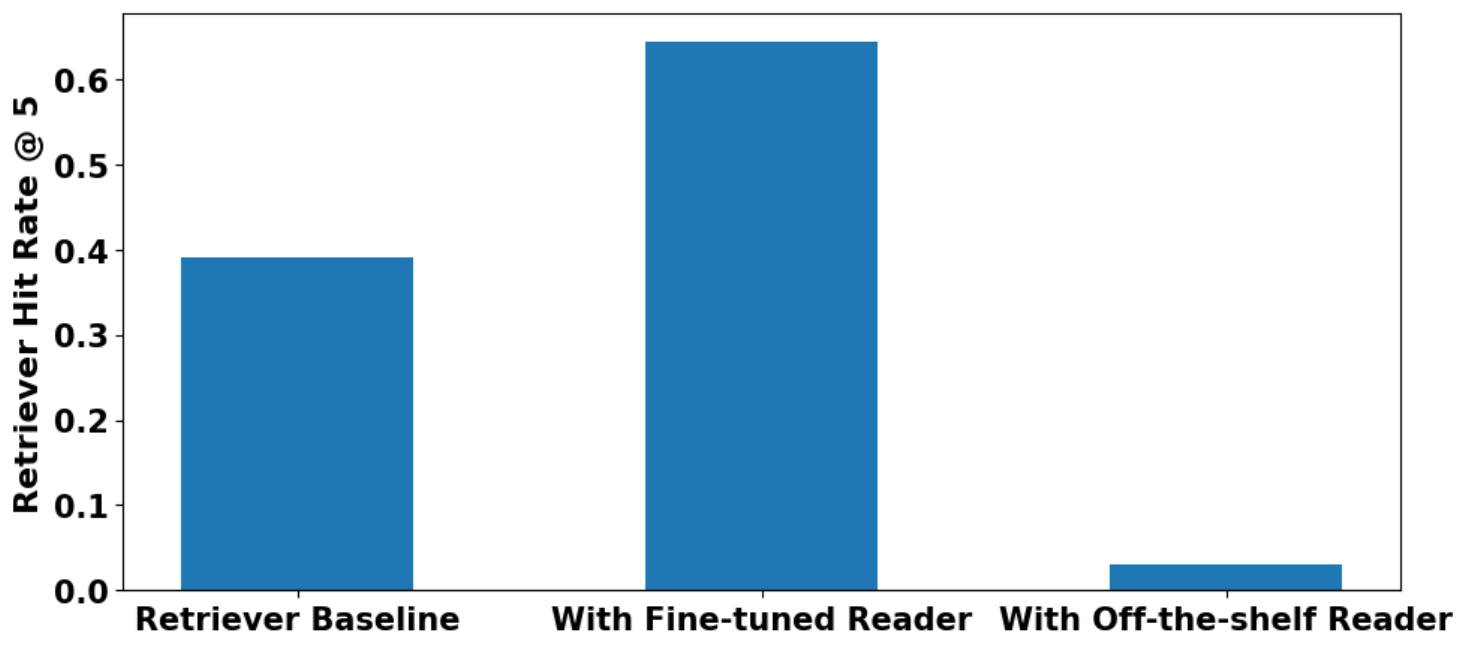} 
    \caption{Training \textit{Contriever} on \textit{NaturalQuestions} for the QA task with attention distillation shows an improved Hit Rate @ 5 with a fine-tuned reader but a significant decline with an off-the-shelf reader.}
    \label{fig:01}
    \vspace{-5mm}  
\end{figure}


\begin{figure*}[t]
\centering
\includegraphics[width=0.9\textwidth]{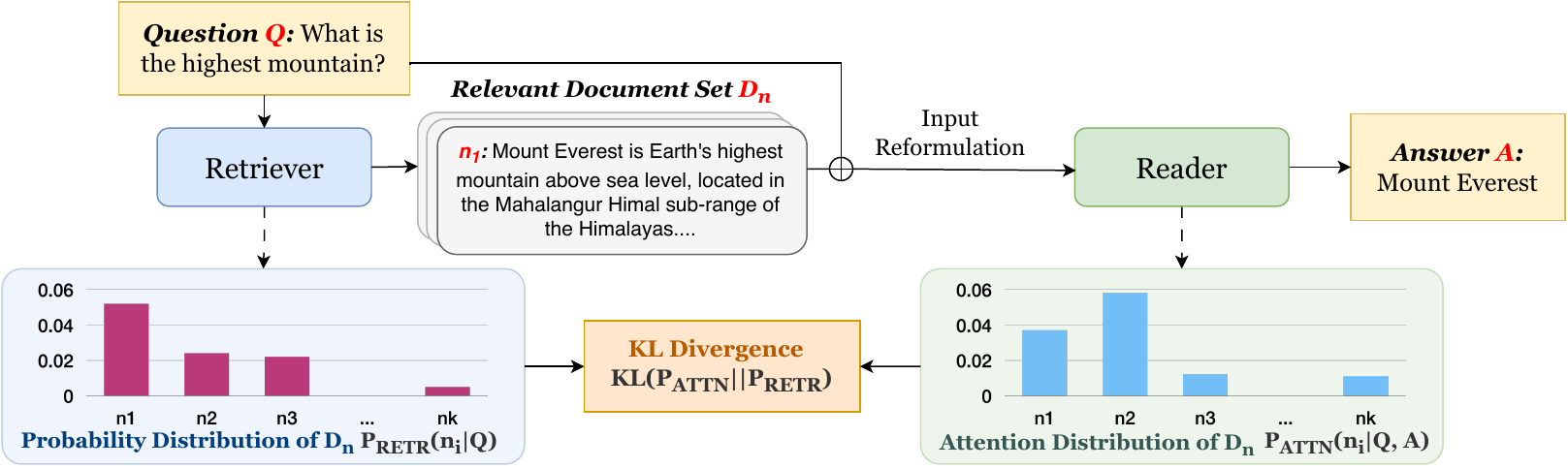} 
\caption{The framework of the Retrieval-augmented Language Model of our experiment.}
\label{fig:02}
\vspace{-4mm}  
\end{figure*}

However, its efficiency heavily relies on the reader model's quality. As Figure \ref{fig:01} shows, low-quality reader models yield ineffective supervision signals, detrimentally impacting the retriever's performance.
A fundamental hypothesis underpinning this mechanism is that more attention to certain tokens suggests greater relevance in answering questions \cite{izacard2020distilling}. Yet this correlation is not clearly defined: no existing work has quantitatively analyzed the role of attention scores within the retrieval-augmented generation framework.
Therefore, our research seeks to understand which text segments gather more attention and how to assess attention quality. 

In this paper, we first experimentally confirmed that attention scores are not always effective as training supervisors in certain experiment settings under QA tasks. Following this, we performed a thorough token-level quantitative analysis to identify commonalities in the distribution of attention scores that align with high-quality supervisory signals. Furthermore, we derived two key indicators to measure distillation quality based on the commonalities. Our main contributions are as follows:


\vspace{-2pt}
\begin{itemize}
    \setlength{\itemsep}{-2pt}
    \item We conduct an extensive analysis of attention scores in language models, mainly focusing on the prevalent decoder-only structure, to understand their impact on retriever model training and the overall performance of retrieval-augmented language models, thereby identifying key factors that significantly influence the model's performance.
    \item We introduce novel metrics to evaluate the reader model's proficiency in attention distillation, aiming to improve training performance by leaning on effective training sessions.
\end{itemize}

\section{Method}
In our experiment, we adapt the ATLAS architecture \cite{izacard2022few} but use a decoder-only language model structure for our empirical analysis, focusing on QA tasks to study attention score distillation mechanisms. 
Specifically, for a given question $Q$, we supply models with a knowledge base $D=\{d_1, d_2, ..., d_m\}$, where each $d_i$ is a unique document. 
The objective of the models is to find the question-relevant documents $D_n=\{n_1, n_2, ..., n_k\} \subseteq D$ using the retriever, and then generate the answer $A$ using the reader.

The attention distillation uses attention scores to gauge the importance of each input document in answer generation. To adapt to changes in the reader's structure, we utilize the \textit{self-attention scores} related to the output tokens as an indicator for indicating document relevance, rather than relying on \textit{cross-attention scores} between input documents and output.
In addition, notice that the contribution of a token $t$ is not only evaluated from the self-attention score $\alpha_t$ but also the norm of the value vector $\textbf{v}_t$ should be taken into account \cite{izacard2022few}. Then the \textit{Softmax} operator is applied to obtain the attention score distribution over the question-relevant documents $D_n$:
\begin{equation}
\setlength\abovedisplayskip{5pt}
\setlength\belowdisplayskip{5pt}
    p_{ATTN}(n_i|Q, A) = Softmax(\sum_{t=1}^T \alpha_t ||\textbf{v}_t||_2)
\end{equation}
where $T$ represents the total number of tokens in $n_i$. 
On the other hand, the retriever's probability distribution $p_{RETR}$ over $D_n$ can be defined as:
\begin{equation}
    p_{RETR}(n_i|Q) = \frac{exp(s(n_i, Q)/\theta)}{\sum_{k=1}^Kexp(s(n_k, Q)/\theta)}
\end{equation}
where $s$ denotes the dot-product of query and document vectors, and $\theta$ is the temperature hyper-parameter.
During training, the attention distribution is distilled into the retriever by minimizing KL-divergence between $p_{ATTN}(n_i)$ and $p_{RETR}(n_i)$.
Figure \ref{fig:02} visually illustrates the retrieval process and the utilization of attention scores during training.


\section{Experiments}
We chose \textit{Falcon-1b} \cite{refinedweb} as our primary decoder-only reader model for its performance and flexibility,
and we follow ATLAS \cite{izacard2022few} in using \textit{Contriver} as the retriever model. 
During the retrieval process, we fix the retrieved documents $D_n$'s size to $k=5$ to balance training costs with the amount of information retrieved, avoiding inefficiencies of either extreme.
\subsection{Experiment Setup}
\textbf{Dataset} We assess the model's performance using the \textit{NaturalQuestions} \cite{kwiatkowski-etal-2019-natural} and the \textit{TriviaQA} \cite{joshi-etal-2017-triviaqa} benchmarks. 
For the knowledge base, we utilize data from Wikipedia as of December 20, 2018. \\
\textbf{Experimental Settings} Specifically, we use the following settings for our experiments. \\
\textbf{1) Off-the-shelf Distillation Training}: We synchronously train the model using the initial \textit{Falcon-1b} \cite{refinedweb} as the reader and \textit{Contriever} \cite{izacard2022unsupervised} as the retriever. \\
\textbf{2) Fine-tuned Distillation Training}: This experiment involves two steps: \\
\textbf{Step1.} We start with the initial \textit{Falcon-1b} as a reader and \textit{Contriever} as a retriever, only fine-tuning the reader while keeping the retriever's parameters fixed. \\
\textbf{Step2.} We continue training the retriever using the fine-tuned reader from Step1, updating the knowledge base index periodically. \\
\textbf{Evaluation Metrics}: We assess the model performance in terms of retrieval quality and question-answering correctness, given the involvement of both retriever and reader models. We use the \textit{top-5} retrieval Hit Rate (HR@5), which is the proportion of retrieved documents $D_n$ containing at least one answer $A$, to measure the retriever's effectiveness. For the reader's QA performance, we employ the standard Exact Match (EM) metric and F1-Score.

\subsection{Results and Discussion}
In this section, we empirically analyze the effectiveness of attention distillation training by answering the following research questions: \\
\textbf{\textit{RQ1: When does the attention distillation work?}}
As shown in Table \ref{fig:01}, the \textit{Fine-tuned Distillation Training} after Step2 shows the best performance in both EM and HR@5. In contrast, \textit{Off-the-shelf Distillation Training} performs the worst, with its retriever even underperforming the initial Contriever model (i.e., the retriever model of \textit{Fine-tuned Distillation Training} Step1). 
Notice that the critical difference lies in the quality of the reader models: \textit{Off-the-shelf Distillation Training} uses the initial Falcon-1b model, whereas \textit{Fine-tuned Distillation Training} employs a well-tuned Falcon-1b. These experimental results strongly suggest that the quality of attention scores is pivotal: \textbf{attention scores from the high-quality readers enhance training, whereas low-quality ones lead to poor interaction between the retriever and the reader.}\\
\textbf{\textit{RQ2: Are there any commonalities in attention scores from the high-quality readers?}}
\setlength{\tabcolsep}{1.5mm}{
\begin{table}[t]\label{tab:performance-drp}
    \centering
    \caption{Model's Performance of Different Experimental Settings}
    \scalebox{1.0}{
    \resizebox{0.5\textwidth}{!}{
    \begin{tabular}{*{6}{c}}
      \toprule
      \multirow{3}*{\textbf{Method}} & \multirow{3}*{\textbf{Dataset}} & \multicolumn{3}{c}{\textbf{Evaluation Metrics}} & \\
      \cmidrule(lr){3-6}
      & & \textbf{EM$\uparrow$} & \textbf{F1$\uparrow$} & \textbf{HR@5$\uparrow$} \\
      \midrule
      Off-the-shelf Distillation & NQ & 27.24 & 33.62 & 0.030 \\
      & TriviaQA & 30.55 & 35.24 & 0.022 \\
      \midrule
      Fine-tuned Distillation (Step1) & NQ & 31.76 & 38.72 & 0.391 & \\
      & TriviaQA & 44.62 & 50.79 & 0.516 \\
      \midrule
      Fine-tuned Distillation (Step2) & NQ & 35.22 & 43.44 & 0.645 & \\
      & TriviaQA & 54.59 & 61.04 & 0.643 \\
      \bottomrule
    \end{tabular}
    }}
    \label{tab:tab01}
    \vspace{-4mm}  
\end{table}}

We sample 1000 data instances from each experiment to obtain reliable analysis results. 
We focus on the attention score characteristics \textbf{at token level} to identify which tokens receive more attention from high-quality signals.
Our analysis firstly finds that in the high-quality readers, the tokens most related to \textit{answer} and \textit{nouns in question} receive the most attention.
Based on our initial observations, we secondly focus on studying the distribution of attention scores for \textit{answer-related} and \textit{question-related} \footnote{We only focus on the nouns in the question in selecting \textit{question-related} tokens.} tokens. We use token embedding's \textit{cosine similarity} to measure its proximity to targets (i.e., answer or nouns in question), 
selecting the top 5\% and top 10\% of closest tokens and analyzing their average \textit{attention scores} and \textit{Spearman correlation with similarity to target tokens}, as shown in Table \ref{tab:02}\footnote{The highest values in the table are highlighted in bold on the NQ Dataset and underlined on the TriviaQA Dataset.}. We also include the \textit{Off-the-shelf Checkpoint} as a baseline to observe attention score evolution in different settings. This analysis identifies the key commonalities in high-quality attention scores.

\setlength{\tabcolsep}{1.5mm}{
\begin{table*}[t]
    \centering
    \caption{Average values of attention scores and Spearman correlation in \textit{answer-related} and \textit{question-related} tokens}
     \scalebox{1.0}{
    \resizebox{0.9\textwidth}{!}{
    \begin{tabular}{*{10}{c}}
      \toprule
      \multirow{3}*{\textbf{Experiment}} & \multirow{3}*{\textbf{Dataset}} & \multicolumn{4}{c}{\textbf{Answer-related}} & \multicolumn{4}{c}{\textbf{Question-related}} \\
      \cmidrule(lr){3-10}
      & & \multicolumn{2}{c}{\textbf{$\boldsymbol{90^{th}}$ percentile}} & \multicolumn{2}{c}{\textbf{$\boldsymbol{95^{th}}$ percentile}} & \multicolumn{2}{c}{\textbf{$\boldsymbol{90^{th}}$ percentile}} & \multicolumn{2}{c}{\textbf{$\boldsymbol{95^{th}}$ percentile}} \\
      \cmidrule(lr){3-10}
      & &  \textbf{Attn.} & \textbf{Corr.} & \textbf{Attn.} & \textbf{Corr.} & \textbf{Attn.} & \textbf{Corr.} & \textbf{Attn.} & \textbf{Corr.} \\
      \midrule
      Off-the-shelf Checkpoint & NQ & 0.033 & 0.227 & 0.039 & 0.196 & 0.023 & 0.103 & 0.024 & 0.092 \\
      \space & TriviaQA & 0.027 & 0.218 & 0.032 & 0.206 & 0.021 & 0.103 & 0.023 & 0.067 \\
       \midrule
      Off-the-shelf Attention Distillation & NQ & 0.017 & 0.145 & 0.017 & 0.076 & 0.027 & 0.139 & 0.039 & 0.153 \\
      \space & TriviaQA & 0.031 & 0.160 & 0.035  & 0.172 & 0.047 & 0.144 & 0.063 & 0.260 \\
       \midrule
      Fine-tuned Attention Distillation (Step1) & NQ & 0.039 & 0.308 & 0.052 & 0.282 & \textbf{0.035} & \textbf{0.343} & \textbf{0.045} & \textbf{0.333} \\
      \space & TriviaQA & 0.058 & 0.259 & 0.074 & 0.258 & 0.058 & 0.349 & \underline{0.078} & \underline{0.372} \\
      Fine-tuned Attention Distillation (Step2) & NQ & \textbf{0.049} & \textbf{0.316} & \textbf{0.066} & \textbf{0.350} & 0.032 & 0.310 & 0.039 & 0.225 \\
      \space & TriviaQA & \underline{0.069} & \underline{0.290} & \underline{0.089} & \underline{0.320} & \underline{0.060} & \underline{0.367} & \underline{0.078} & 0.326 \\
      \bottomrule
    \end{tabular}
    \vspace{-4pt}
    }
}
    \label{tab:02}
    \vspace{-4pt}
\end{table*}
\vspace{-2pt}
}

\begin{figure*}[htbp!]
    \centering
    \includegraphics[width=0.95\textwidth]{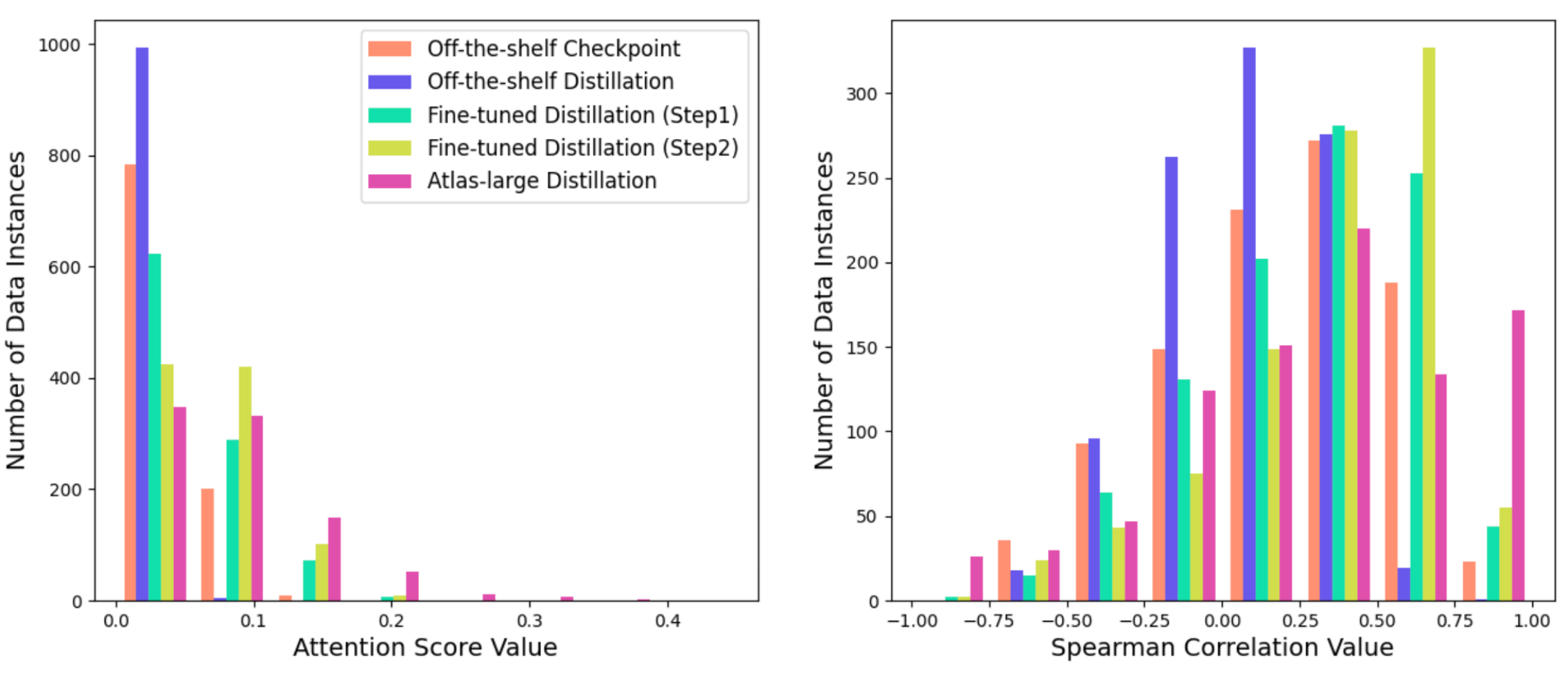}
    \caption{The attention score distribution histogram (left) and Spearman correlation distribution histogram of $95^{th}$ percentile \textit{answer-related} tokens under NQ dataset.}
    \label{fig:05}
\end{figure*}

\textbf{Commonality1. Higher attention to answer tokens in higher-quality models.} 
In all training settings, tokens closer to answer tokens  (i.e., from a similarity higher than $90^{th}$ percentile to a similarity higher than $95^{th}$ percentile) receive increasingly higher attention scores.
It can be observed that for both two measure metrics, 
the \textit{Off-the-shelf Distillation Training} results are lower compared to the \textit{Off-the-shelf Checkpoint}, while \textit{Fine-tuned Distillation Training} shows improvement in both Step1 and Step2. 
The results suggest that in \textit{Off-the-shelf Distillation}, the reader's attention does not effectively "highlight" key information, leading to suboptimal training. In contrast, \textit{Fine-tuned Distillation} after Step1 and Step2 both indicate that high-quality readers focus more on relevant answer tokens, thereby enhancing both the retriever's performance and the relevance of attention allocated to these tokens. We can see more details according to Figure \ref{fig:05}.



\textbf{Commonality 2. Tokens similar to question nouns receive more attention in high-quality models.} 
Table \ref{tab:02} also indicates that tokens closer to the nouns in question tokens receive higher attention scores.
The \textit{Fine-tuned Distillation} experiments exhibit much higher values in both metrics compared to \textit{Off-the-shelf Checkpoint} and \textit{Off-the-shelf Attention Distillation}, aligning with their superior performance. 
However, unlike Commonality 1, the Spearman correlation between attention to question-related tokens and model performance isn't consistent: while \textit{Fine-tuned Attention Distillation} Step2 surpasses Step1, its metric values do not consistently align with this improvement, suggesting a more complex relationship. \\
\textbf{\textit{RQ3: How do we evaluate the quality of attention distillation on decoder-only readers based on the analysis results?}} 

\begin{figure}
    \centering
    \includegraphics[width=0.5\textwidth]{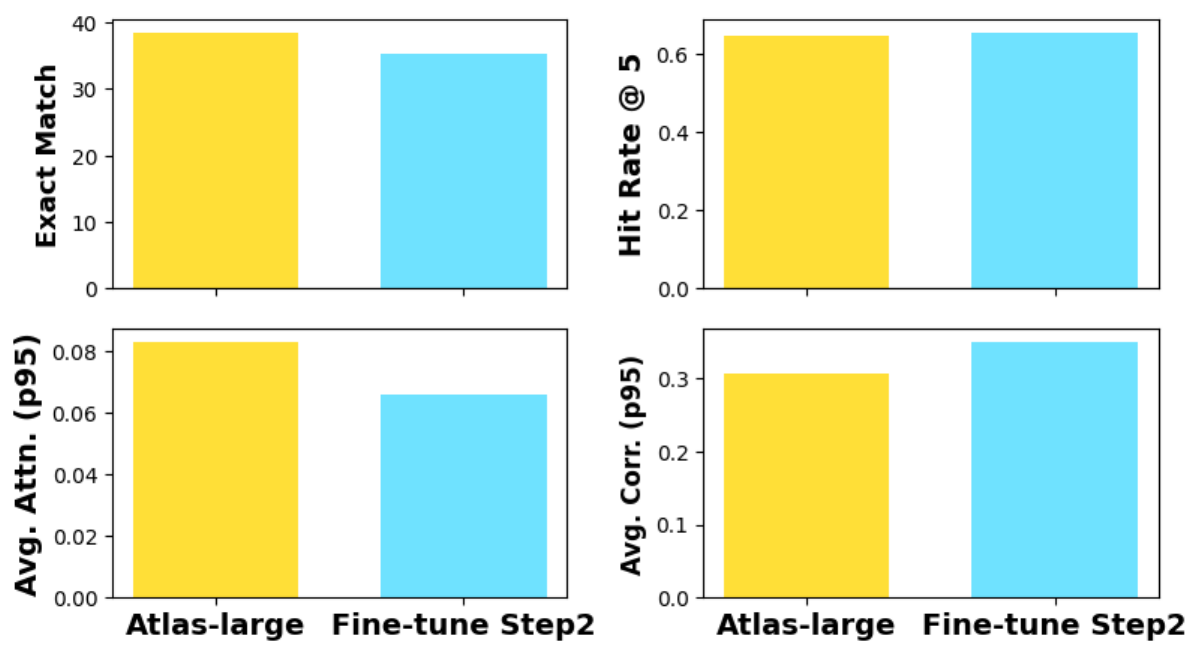}
    \caption{Model performance (top) and their attention distillation analysis (bottom) of \textit{Atlas-large} model (yellow) for the answer-related tokens, comparing with \textit{Fine-tuned Distillation Training (Step2)} (blue).}
    \label{fig:04}
    \vspace{-12pt}
\end{figure}

\textbf{Indicator1.} Focusing on the attention scores of the nearest tokens to answer $A$, denoted as $M_A=\{ma_1, ..., ma_k\}$. Higher average $P_{ATTN}(ma_i)$ values indicate better attention distillation quality. Additionally, a higher average Spearman correlation between the $P_{ATTN}(ma_i)$ and their semantic similarity to $A$ also signifies better quality.

\textbf{Indicator2.} Examining the attention scores of tokens closest to nouns in question $Q$, denoted as $M_Q=\{mq_1, ..., mq_k\}$. An increase in average $P_{ATTN}(mq_i)$ suggests better quality. Moreover, if the average Spearman correlation between the attention scores of $M_Q$ and their similarity to $Q$ is above the threshold for a weak monotonic relationship (i.e., value > 0.3), the attention distillation quality is considered good. \\
\textbf{\textit{RQ4: Can we extend the proposed indicators to encoder-to-decoder structure readers?}}

An analysis with the fine-tuned encoder-to-decoder structure \textit{Atlas-large} model is presented in Figure \ref{fig:04}.
The results show that the performance of \textit{Atlas-large} surpasses \textit{Fine-tuned Distillation Training (Step2)}. However, only the average $P_{ATTN}(ma_i)$ trend from Indicator1 applies to this encoder-to-decoder structure model, while \textit{Atlas-large} exhibits a polarized distribution for the Spearman correlation values. (see Appendix \ref{sec:appendix A}).\\
\textbf{\textit{RQ5: Can we extend the proposed indicators to perplexity distillation training?}}




Finally, we want to determine if our indicators can apply to perplexity distillation, another popular knowledge distillation method used in training the retriever model. We fine-tune \textit{Atlas-large} model with the perplexity distillation method and find that the perplexity distribution does not align with either Commonality 1 or Commonality 2, saying that our indicators are not suitable for perplexity distillation (details in Appendix \ref{sec:appendix A} and \ref{sec:appendix B}).

\section{Related Work}
The concept of using attention scores for knowledge distillation was introduced by \cite{izacard2020distilling}, and the following research has mainly focused on independently optimizing the reader (i.e., the large language model) and the retriever. Previous studies have explored improving large language model performance within the retriever-then-read framework by addressing issues like hallucination \cite{shuster2021retrieval} and dependency on pre-training data \cite{kandpal2023large}, and enhancing retriever efficiency through techniques like specific data sampling \cite{hofstatter2021efficiently} or using large language model generators as retrievers \cite{yu2022generate}. Only one study has examined the reader-retriever interaction within a neural-retrieval-in-the-loop architecture, noting that imperfect retrievers can degrade reader performance, though it lacked quantitative analysis \cite{behnamghader2022can}.

Unlike the previous work, our paper conducts a comprehensive quantitative analysis of how the reader and the retriever interact during the neural-retrieval-in-the-loop training architecture under the attention distillation mechanism. We propose novel metrics to identify whether the training process is beneficial, which can apply to all general neural-retrieval-in-the-loop architecture.

\section{Conclusion}
In this paper, we comprehensively evaluate attention distillation for training retrieval-augmented language models, emphasizing the importance of attention to answer and question-related tokens. We further introduce novel metrics for assessing language models' attention distillation ability to optimize the training process.

\section{Limitation}
This paper analyzes the attention score-based knowledge distillation quality in training retrieval-augmented language models under various experimental settings in QA tasks. Furthermore, based on our findings, we have developed two indicators to assess the quality of attention score supervision. However, our exploration is conducted based on lightweight language models (i.e., language models with about one billion parameters) due to their flexibility and have yet to extend to larger-scale language models. In future work, we will focus on validating the accuracy of our methods on more extensive language models to enhance the generalizability and applicability of our results.

\bibliography{anthology,custom}

\appendix

\clearpage
\section{Quantitative Analysis of Answer-Related Tokens}
\label{sec:appendix A}
\begin{figure*}[htbp!]
    \centering
    \includegraphics[width=1.0\textwidth]{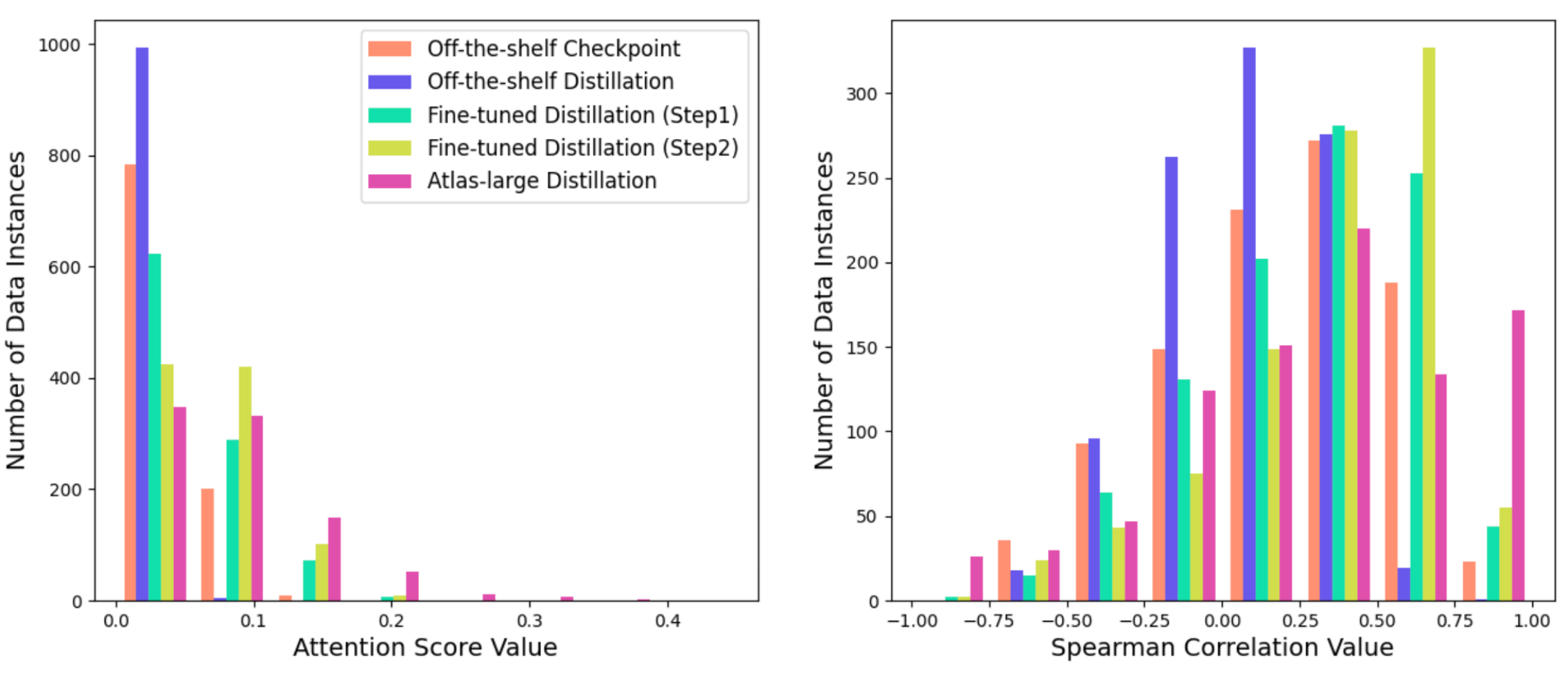} 
    \caption{The attention score distribution histogram (left) and Spearman correlation distribution histogram of $95^{th}$ percentile \textit{answer-related} tokens under NQ dataset.}
    \label{fig:a01}
\end{figure*}

\begin{figure*}[htbp!]
    \centering
    \includegraphics[width=1.0\textwidth]{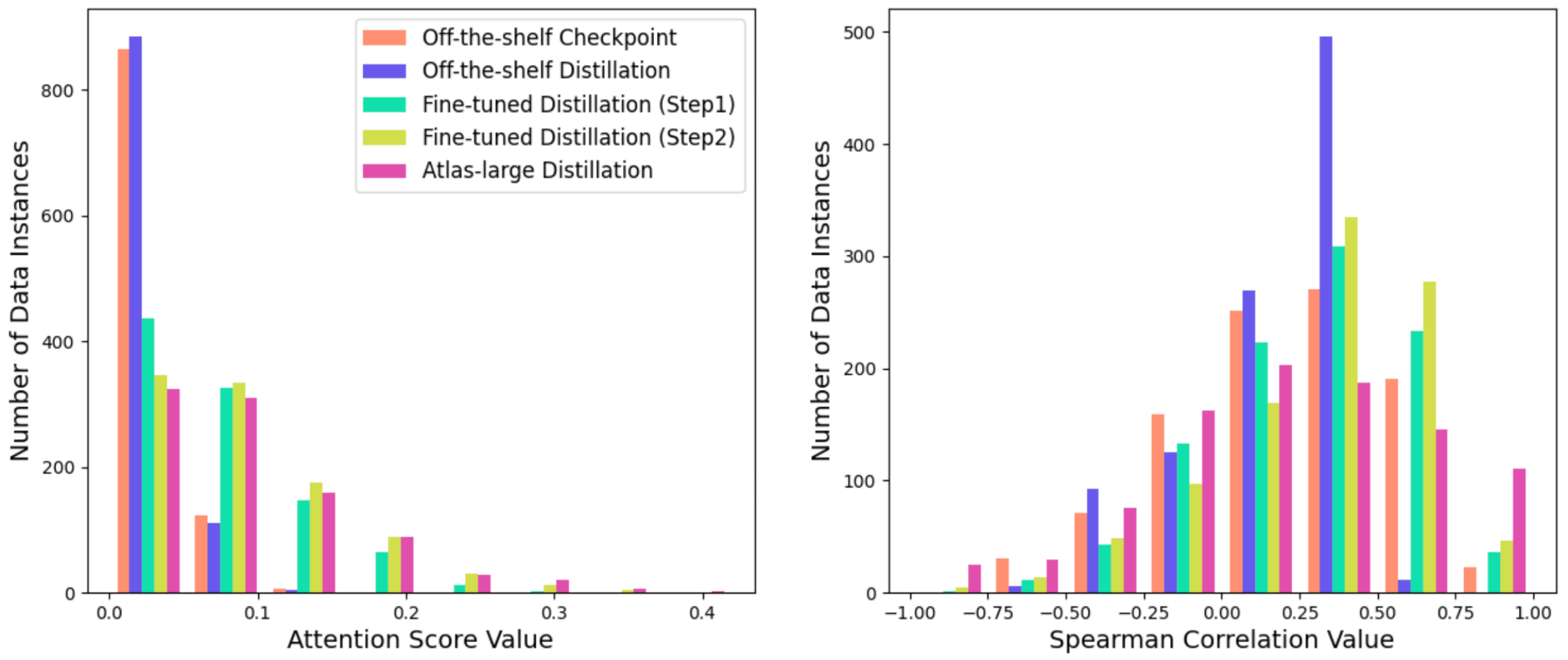} 
    \caption{The attention score distribution histogram (left) and Spearman correlation distribution histogram of $95^{th}$ percentile \textit{answer-related} tokens under TriviaQA dataset.}
    \label{fig:a02}
\end{figure*}

We present a detailed analysis of \textit{answer-related} tokens' attention score distribution (or perplexity distribution of \textit{Perplexity Distillation Training}) shown in Table \ref{tab:04}, Figure \ref{fig:a01}, and Figure \ref{fig:a02}.

\setlength{\tabcolsep}{1.5mm}{
    \begin{table*}[htbp!]
        \centering
        \caption{Mean and std. of attention scores (or perplexity distribution in \textit{Perplexity Distillation Training}) and the Spearman correlations of the answer-related tokens.}
         \scalebox{1}{
        \resizebox{0.98\textwidth}{!}{
        \begin{tabular}{*{9}{c}}
          \toprule
          \textbf{Experiment} & \textbf{Dataset} & \textbf{Avg. Attn. (p90)} & \textbf{Spearman Corr. (p90)} & \textbf{Avg. Attn. (p95)} & \textbf{Spearman Corr. (p95)} &\\
          \midrule
          Off-the-shelf Model Checkpoint & NQ & $0.033 \pm 0.016$ & $0.227 \pm 0.259$ & $0.039 \pm 0.023$ & $0.196 \pm 0.349$ \\
           & TriviaQA & $0.027 \pm 0.013$ & $0.218 \pm 0.252$ & $0.032 \pm 0.019$ & $0.206 \pm 0.331$ \\
           \midrule
          Off-the-shelf Attention Distillation & NQ & $0.017 \pm 0.008$ & $0.145 \pm 0.193$ & $0.017 \pm 0.010$ & $0.076 \pm 0.254$ \\
           & TriviaQA & $0.031 \pm 0.012$ & $0.160 \pm 0.174$ & $0.035 \pm 0.017$ & $0.172 \pm 0.236$ \\
           \midrule
          Fine-tuned Distillation Training (Step1) & NQ & $0.039 \pm 0.023$ & $0.308 \pm 0.276$ & $0.052 \pm 0.036$ & $0.282 \pm 0.336$ \\
           & TriviaQA & $0.058 \pm 0.031$ & $0.259 \pm 0.261$ & $0.074 \pm 0.050$ & $0.258 \pm 0.331$ \\
          Fine-tuned Distillation Training (Step2) & NQ & $0.049 \pm 0.023$ & $0.316 \pm 0.280$ & $0.066 \pm 0.036$ & $0.350 \pm 0.336$ \\
           & TriviaQA & $0.069 \pm 0.036$ & $0.290 \pm 0.267$ & $0.089 \pm 0.061$ & $0.320 \pm 0.323$ \\
          \midrule
          Atlas-large Distillation Training & NQ & $0.062 \pm 0.036$ & $0.171 \pm 0.462$ & $0.083 \pm 0.058$ & $0.307 \pm 0.471$ \\
          & TriviaQA & $0.072 \pm 0.045$ & $0.141 \pm 0.379$ & $0.091 \pm 0.067$ & $0.217 \pm 0.438$ \\
          \midrule
          Perplexity Distillation Training & TriviaQA & $0.072 \pm 0.039$ & $0.029 \pm 0.142$ & $0.071 \pm 0.042$ & $0.013 \pm 0.202$ \\
          \bottomrule
        \end{tabular}
        }
    }
    \label{tab:04}
    \end{table*}
}

\section{Quantitative Analysis of Question-Related Tokens}
\label{sec:appendix B}
\begin{figure*}[htbp!]
    \centering
    \includegraphics[width=1.0\textwidth]{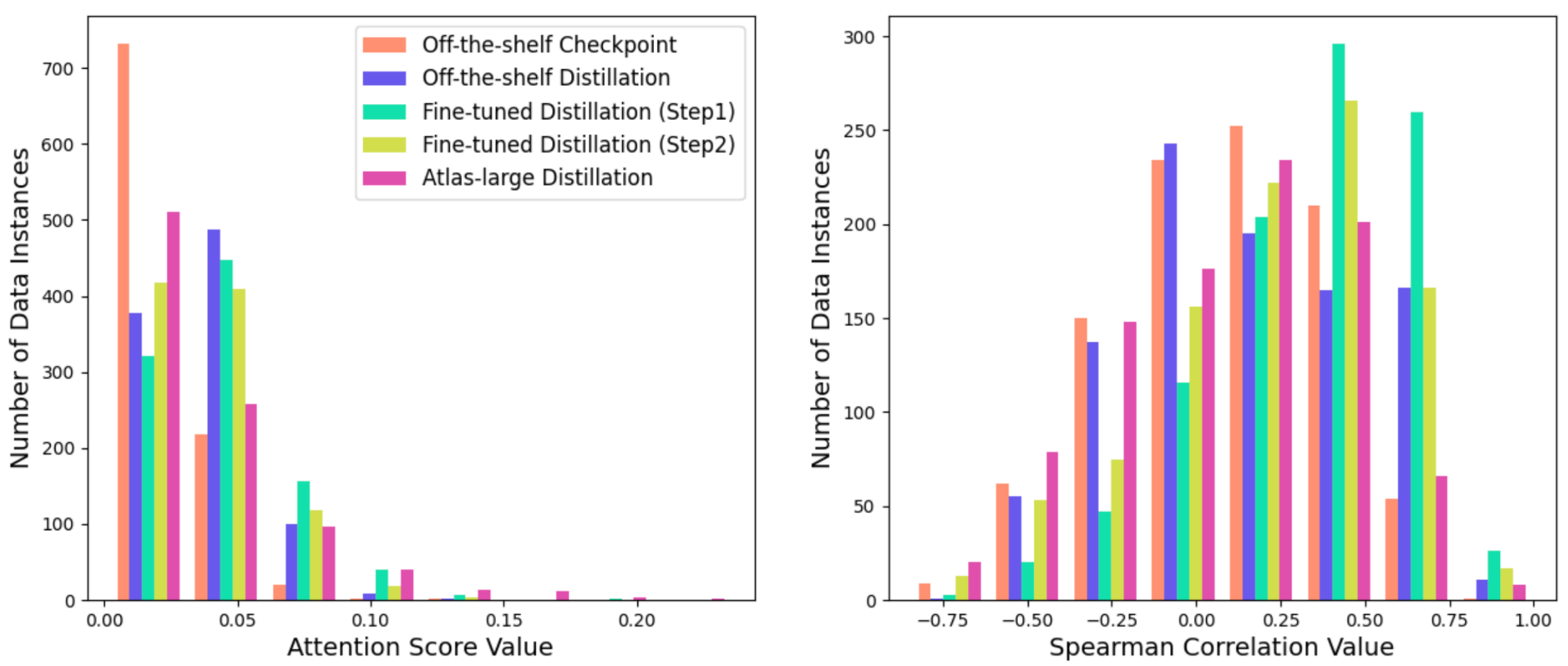} 
    \caption{The attention score distribution histogram (left) and Spearman correlation distribution histogram of $95^{th}$ percentile \textit{question-related} tokens under NQ dataset.}
    \label{fig:a03}
\end{figure*}

\begin{figure*}[htbp!]
    \centering
    \includegraphics[width=1.0\textwidth]{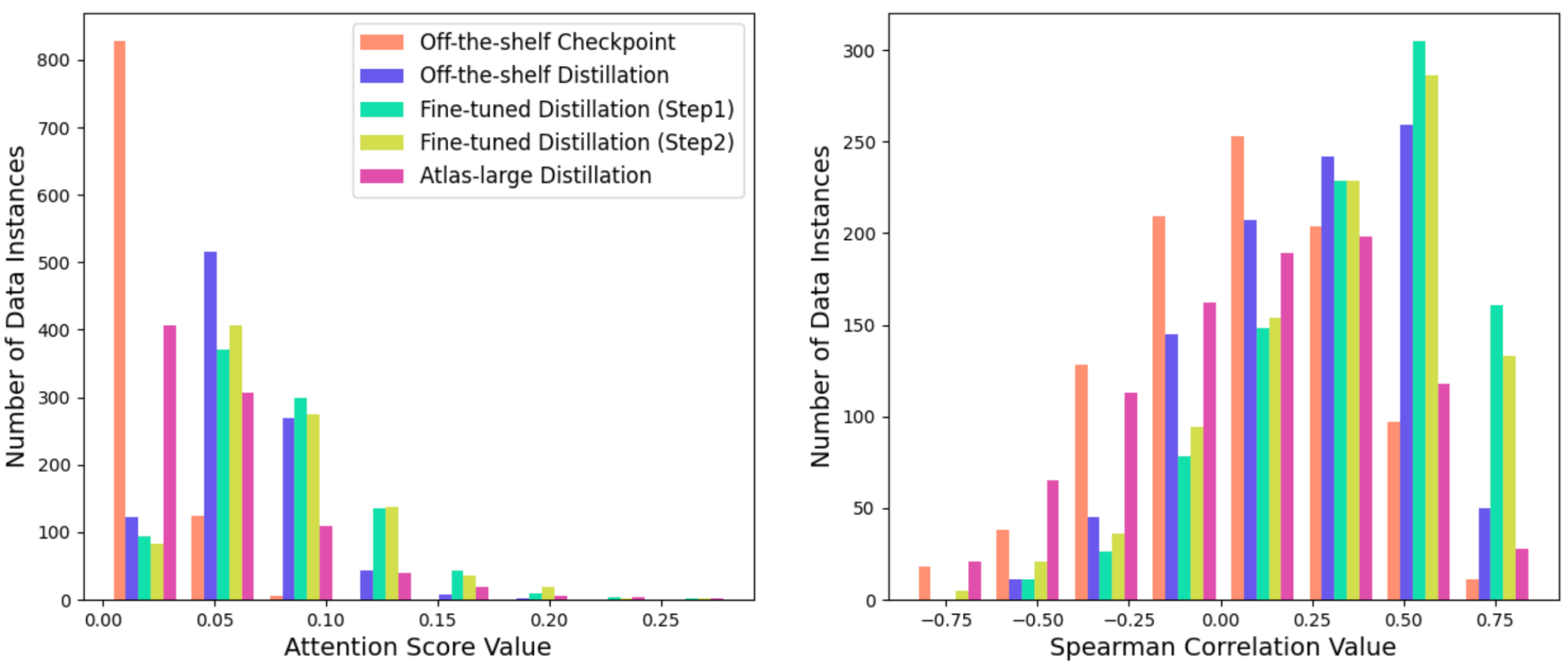} 
    \caption{The attention score distribution histogram (left) and Spearman correlation distribution histogram of $95^{th}$ percentile \textit{question-related} tokens under the TriviaQA dataset.}
    \label{fig:a04}
\end{figure*}

We present a detailed analysis of \textit{question-related} tokens' attention score distribution (or perplexity distribution of \textit{Perplexity Distillation Training}) shown in Table \ref{tab:05}, Figure \ref{fig:a03}, and Figure \ref{fig:a04}.

\setlength{\tabcolsep}{1.5mm}{
\begin{table*}[htbp!]
    \centering
    \caption{Mean and std. of average attention scores (or perplexity distribution in \textit{Perplexity Distillation Training}) and Spearman correlations of the question-related tokens}
     \scalebox{1}{
    \resizebox{1.0\textwidth}{!}{
    \begin{tabular}{*{9}{c}}
      \toprule
      \textbf{Experiment} & \textbf{Dataset} & \textbf{Avg. Attn. (p90)} & \textbf{Spearman Corr. (p90)} & \textbf{Avg. Attn. (p95)} & \textbf{Spearman Corr. (p95)} &\\
      \midrule
      Off-the-shelf Model Checkpoint & NQ & $0.023 \pm 0.011$ & $0.103 \pm 0.253$ & $0.024 \pm 0.014$ & $0.092 \pm 0.309$ \\
       & TriviaQA & $0.021 \pm 0.010$ & $0.103 \pm 0.245$ & $0.023 \pm 0.013$ & $0.067 \pm 0.304$ \\
       \midrule
      Off-the-shelf Attention Distillation & NQ & $0.027 \pm 0.010$ & $0.139 \pm 0.237$ & $0.039 \pm 0.017$ & $0.153 \pm 0.341$ \\
       & TriviaQA & $0.047 \pm 0.016$ & $0.144 \pm 0.220$ & $0.063 \pm 0.025$ & $0.260 \pm 0.280$ \\
       \midrule
      Fine-tuned Distillation Training (Step1) & NQ & $0.035 \pm 0.015$ & $0.343 \pm 0.238$ & $0.045 \pm 0.023$ & $0.333 \pm 0.303$ \\
       & TriviaQA & $0.058 \pm 0.024$ & $0.349 \pm 0.222$ & $0.078 \pm 0.037$ & $0.372 \pm 0.285$ \\
      Fine-tuned Distillation Training (Step2) & NQ & $0.032 \pm 0.014$ & $0.310 \pm 0.256$ & $0.039 \pm 0.021$ & $0.225 \pm 0.340$ \\
       & TriviaQA & $0.060 \pm 0.025$ & $0.367 \pm 0.227$ & $0.078 \pm 0.037$ & $0.326 \pm 0.311$ \\
      \midrule
      Atlas-large Distillation Training & NQ & $0.037 \pm 0.027$ & $0.082 \pm 0.251$ & $0.038 \pm 0.032$ & $0.086 \pm 0.345$ \\
      & TriviaQA & $0.047 \pm 0.245$ & $0.076 \pm 0.249$ & $0.050 \pm 0.038$ & $0.081 \pm 0.348$ \\
      \midrule
      Perplexity Distillation Training & TriviaQA & $0.063 \pm 0.038$ & $-0.012 \pm 0.207$ & $0.060 \pm 0.042$ & $-0.036 \pm 0.297$ \\
      \bottomrule
    \end{tabular}
    }}
    \label{tab:05}
\end{table*}
}

\section{Dataset Statistics}
\label{sec:appendix C}
For the \textit{NaturalQuestions} dataset, we split it according to the number of 79,168/8,757/3,610 to form the train/validation/test dataset; for the \textit{TriviaQA} dataset, we split it according to the number of 78,785/8,837/11,313 to form the train/validation/test dataset.

\section{Implementation Details}
\label{sec:appendix D}

We conducted all computations on a Nvidia A100 GPU. For the \textit{Off-the-shelf Distillation Training} and the \textit{Fine-tuned Distillation Training}, we use \textit{Falcon-1b} as the initial reader model and \textit{Contriever} as the initial retriever model, which has about 1 billion and 110 million training parameters respectively. For the \textit{Atlas-large Distillation Training} and \textit{Perplexity Distillation Training}, we use \textit{T5-large} as the initial reader model and \textit{Contriever} as the initial retriever model, which has about 770 million and 110 million training parameters respectively. \\
\textbf{Off-the-shelf Distillation Training}
We set the batch size to 1, the maximum length of the input prompt to 128, and limit the generation max length to 32. We set the learning rate to 1e-5 and used the Adam optimizer. 
For \textit{NaturalQuestions} dataset, we set the total training steps to 160,000 with approximately 2000 warmup steps, training for about 40 hours. 
For \textit{TriviaQA} dataset, we set the total training steps to 320,000 with approximately 4000 warmup steps, training for about 60 hours. 

\noindent
\textbf{Fine-tuned Distillation Training}
For Step 1, we set the batch size to 1, the maximum length of the input prompt to 128, and limit the generation max length to 32. 
We set the learning rate to 1e-5 and used the Adam optimizer.
For \textit{NaturalQuestions} dataset, we set the total training steps to 160,000 with approximately 2000 warmup steps, training for about 30 hours. 
For \textit{TriviaQA} dataset, we set the total training steps to 320,000 with approximately 4000 warmup steps, training for about 45 hours. 

For Step 2, we set the batch size to 1, the maximum length of the input prompt to 128, and limit the generation max length to 32.
We set the learning rate to 5e-7 and used the Adam optimizer.
For \textit{NaturalQuestions} dataset, we set the total training steps to 6,000 with approximately 300 warmup steps, training for about 2 hours.
For \textit{TriviaQA} dataset, we set the total training steps to 32,000 with approximately 600 warmup steps, training for about 3 hours. \\
\textbf{Atlas-large Distillation Training} 
We set the batch size to 1, the maximum length of the input prompt to 128, and limit the generation max length to 32. We set the learning rate to 4e-5 and used the Adam optimizer. 
For \textit{NaturalQuestions} dataset, we set the total training steps to 10,000 with approximately 500 warmup steps, training for about 20 hours. 
For \textit{TriviaQA} dataset, we set the total training steps to 30,000 with approximately 600 warmup steps, training for about 40 hours. \\
\textbf{Perplexity Distillation Training} 
We set the batch size to 1, the maximum length of the input prompt to 128, and limit the generation max length to 32. We set the learning rate to 4e-5 and used the Adam optimizer. 
For \textit{NaturalQuestions} dataset, we set the total training steps to 20,000 with approximately 1000 warmup steps, training for about 40 hours. 
For \textit{TriviaQA} dataset, we set the total training steps to 10,000 with approximately 500 warmup steps, training for about 15 hours.


\end{document}